\documentclass{article} 
\usepackage{iclr2024_conference,times}
\usepackage{makecell}
\usepackage{tabularx}

\usepackage{amsmath,amsfonts,bm}

\def\eqref#1{equation~\ref{#1}}

\def\1{\bm{1}}

\DeclareMathAlphabet{\mathsfit}{\encodingdefault}{\sfdefault}{m}{sl}
\SetMathAlphabet{\mathsfit}{bold}{\encodingdefault}{\sfdefault}{bx}{n}

\usepackage{setspace}
\usepackage[hidelinks]{hyperref}
\usepackage{fontawesome}
\usepackage{xcolor}

\hypersetup{
    colorlinks,
    linkcolor={magenta},
    citecolor={cyan},
    urlcolor={magenta}
}

\usepackage{url}
\usepackage{colortbl}

\usepackage{amsmath,bm}
\usepackage{graphicx}
\usepackage{microtype}
\usepackage{amsfonts}
\usepackage[noend]{algpseudocode}
\usepackage{algorithmicx,algorithm}
\usepackage{color}
\usepackage{etoolbox,lipsum}
\usepackage{float}
\usepackage{caption}
\usepackage{booktabs}
\usepackage{multirow}
\usepackage{array}
\usepackage{xspace}
\usepackage{enumerate}
\usepackage{nth}
\usepackage{cleveref}
\usepackage{bbm}
\usepackage{longtable}
\usepackage{CJKutf8}
\usepackage{tcolorbox}
\usepackage{xcolor}
\tcbuselibrary{breakable}

\title{Technical Report of TeleChat2, TeleChat2.5 and T1}

\begin{document}

\maketitle

\begin{abstract}

We introduce the latest series of TeleChat models: \textbf{TeleChat2}, \textbf{TeleChat2.5}, and \textbf{T1}, offering a significant upgrade over their predecessor, TeleChat. Despite minimal changes to the model architecture, the new series achieves substantial performance gains through enhanced training strategies in both pre-training and post-training stages. The series begins with \textbf{TeleChat2}, which undergoes pretraining on 10 trillion high-quality and diverse tokens. This is followed by Supervised Fine-Tuning (SFT) and Direct Preference Optimization (DPO) to further enhance its capabilities. \textbf{TeleChat2.5} and \textbf{T1} expand the pipeline by incorporating a continual pretraining phase with domain-specific datasets, combined with reinforcement learning (RL) to improve performance in code generation and mathematical reasoning tasks. The \textbf{T1} variant is designed for complex reasoning, supporting long Chain-of-Thought (CoT) reasoning and demonstrating substantial improvements in mathematics and coding. In contrast, \textbf{TeleChat2.5} prioritizes speed, delivering rapid inference. Both flagship models of \textbf{T1} and \textbf{TeleChat2.5} are dense Transformer-based architectures with 115B parameters, showcasing significant advancements in reasoning and general task performance compared to the original TeleChat. Notably, \textbf{T1-115B} outperform proprietary models such as OpenAI's o1-mini and GPT-4o. We publicly release \textbf{TeleChat2}, \textbf{TeleChat2.5} and \textbf{T1}, including post-trained versions with 35B and 115B parameters, to empower developers and researchers with state-of-the-art language models tailored for diverse applications.

\end{abstract}

\begin{table}[htbp] \small
\begin{center}
\renewcommand{\arraystretch}{1.3}
\begin{tabular}{ll}
\toprule
     \textbf{Model} & \textbf{Link}  \\ 
\midrule
\rowcolor{pink!25}\textbf{TeleChat2-35B} & \url{https://modelscope.cn/models/TeleAI/TeleChat2-35B}  \\
 \textbf{TeleChat2-115B} & \url{https://modelscope.cn/models/TeleAI/TeleChat2-115B}  \\
 \rowcolor{pink!25}\textbf{TeleChat2.5-35B} & \url{https://modelscope.cn/models/TeleAI/TeleChat2.5-35B}  \\
\textbf{TeleChat2.5-115B} & \url{https://modelscope.cn/models/TeleAI/TeleChat2.5-115B}  \\
\rowcolor{pink!25}\textbf{T1-35B} & \url{https://modelscope.cn/models/TeleAI/T1-35B}  \\
\textbf{T1-115B} & \url{https://modelscope.cn/models/TeleAI/T1-115B}  \\

\bottomrule
\end{tabular}
\end{center}

\vspace{-2mm}
\end{table}

\begin{table}[htbp] \small
\begin{center}
\renewcommand{\arraystretch}{1.3}
\begin{tabular}{ll}
\toprule
     \textbf{Model Series} & \textbf{Github Link}  \\ 
\midrule
 \rowcolor{pink!25}\textbf{TeleChat2} &   \url{https://github.com/Tele-AI/TeleChat2}  \\
\textbf{TeleChat2.5} & \url{https://github.com/Tele-AI/TeleChat2.5} \\
\rowcolor{pink!25}\textbf{T1} & \url{https://github.com/Tele-AI/T1} \\

\bottomrule
\end{tabular}
\end{center}

\vspace{-2mm}
\end{table}

\newpage
\tableofcontents

\section{Introduction}
\label{intro}

In recent years, there has been swift advancement and growth in Large Language Models (LLMs), indicating strides toward Artificial General Intelligence (AGI). Proprietary products such as GPT4 \citep{openai2024gpt4technicalreport}, Claude \citep{claude}, and Gemini\citep{gemini} demonstrate performance at par with human capabilities, elevating the community's expectations for the potential of open-source LLMs. In addition to proprietary models, several notable open-source LLMs, such as LLaMA series (\cite{llama}; \cite{llama2}; \cite{grattafiori2024llama3herdmodels}), Qwen series(\cite{bai2023qwen}; \cite{yang2024qwen2technicalreport}; \cite{qwen2025qwen25technicalreport}), Mistral series(\cite{jiang2023mistral7b};\cite{jiang2024mixtralexperts}), Deepseek series(\cite{deepseekai2024deepseekv2strongeconomicalefficient}; \cite{deepseekai2024deepseekv3technicalreport}), and our TeleChat series (\cite{telechat}) are also making significant progress, striving to narrow the gap with their proprietary counterparts. The open-weight models have made large language models accessible to developers, enabling wider participation in research, promoting innovation through community collaboration, and speeding up the development of AI applications across various fields. Recent advancements, such as the success of DeepSeek-R1 \citep{deepseekai2025deepseekr1incentivizingreasoningcapability}, demonstrate the critical role of long Chain-of-Thought (COT) and reinforcement learning (RL) in enhancing the reasoning capabilities of large language models (LLMs). Notable examples, including OpenAI-o1 \citep{openai2024openaio1card}, Skywork OR1 \citep{he2025skyworkopenreasoner1}, Qwen3 \citep{yang2025qwen3technicalreport}, and Kimi-K1.5 \citep{kimiteam2025kimik15scalingreinforcement}, exemplify how RL can significantly improve performance in complex tasks such as mathematical reasoning and code generation.

To advance open-source contributions, we have enhanced our models and introduced the latest series including \textbf{TeleChat2}, \textbf{TeleChat2.5}, and \textbf{T1}, representing a significant upgrade over their predecessor, TeleChat. The open-weight releases include post-trained variants of 35B and 115B parameter language models. We have made the model parameters publicly available on platforms such as HuggingFace and ModelScope. Additionally, we provide the full codebase on GitHub, which includes comprehensive tools for model fine-tuning, quantization, deployment, and integration with LangChain, enabling a wide range of practical applications.

The development of the new series of TeleChat consists of two main stages: (1) A pre-training stage in which the model is trained on massive datasets by predicting the next word in continuous text. 
The pre-training process of TeleChat2 is meticulously detailed, highlighting the preparation of diverse data types and data composition adjustment. TeleChat2 efficiently captures long-term dependencies, initially trained on 8K tokens before advancing to 256K tokens in the pre-training stages, exhibiting remarkable performance on long-context benchmarks. (2) Following this, we conduct post-training, including Supervised Fine-Tuning (SFT, \cite{ouyang2022training}) and Direct Preference Optimization (DPO, \cite{dpo2023}) on the base model of TeleChat2, to align it with human preferences and further improve specific capabilities. This paper also present our insights into enhancing specific capabilities such as code, reasoning, tool use, and precise instruction following. 

The \textbf{TeleChat2}, \textbf{TeleChat2.5}, and \textbf{T1} series were trained on the Atlas 900 A2 cluster, powered by 8,000 Ascend NPUs. Distributed training leverages the 4D parallelism strategy provided by MindSpore’s large-model parallel framework\footnote{\url{https://www.mindspore.cn/}}, enabling efficient scaling for trillion-parameter models. The computational infrastructure was hosted at CTYun’s Shanghai Compute Center, which delivered the high-performance resources required for large-scale training.

The new series is developed to enhance the model’s ability to understand and generate natural language text, particularly in complex and nuanced contexts. To evaluate their performance in these scenarios, we test \textbf{TeleChat2}, \textbf{TeleChat2.5}, and \textbf{T1} across a comprehensive set of benchmarks spanning mathematics, reasoning, tool usage, precise instruction following, and open-ended tasks. Evaluation results demonstrate that these models achieve significant advancements in reasoning and general task performance compared to there predecessor, TeleChat. Notably, \textbf{T1-115B} outperforms proprietary models like \texttt{OpenAI’s o1-mini} and \texttt{GPT-4o}.

\begin{itemize}
\item \textbf{Better in training data.} We improve both the quality and quantity of the data used for pre-training and post-training. During the pretraining stage, we expand the high-quality pre-training datasets from the previous 3 trillion tokens to 10 trillion tokens, laying a robust groundwork for common sense, expert knowledge, and reasoning capabilities. As for the post-training data, we employ more rigorous quality control and filtering process, and meticulously gather high-quality data to enhance several specific capabilities. Additionally, we conduct data blending experiments to identify the optimal data composition for both pre-training and post-training phases.

\item \textbf{Better in model size.} We develop the new model series at a significantly larger scale compared to previous iterations of the TeleChat series. Our flagship language models of \textbf{TeleChat2}, \textbf{TeleChat2.5} and \textbf{T1} feature 115 billion trainable parameters. Furthermore, we also introduce 35B variants, providing a more cost-effective solution for resource-constrained scenarios. Given that the new series of models share a uniform model architecture and are trained on the homogeneous source data but in varying sizes, they can collaborate under the AI-Flow framework \citep{shao2024aiflownetworkedge} \citep{an2025aiflowperspectivesscenarios}, distributing the workload across multiple models situated in diverse computational nodes, including end devices, edge nodes, and cloud servers. This facilitates a seamless flow of intelligence across networks.

\item \textbf{Better in real-life applications.} The new model series are trained to significantly extend the model's contextual length beyond that of TeleChat, supporting context window up to 128K tokens. This enhancement is essential for real-life applications such as lengthy conversations, long-distance reasoning and understanding, and other tasks that require the model to consider a substantial amount of context. Additionally, the new series also provides better and easier tool usage, making it more accessible and user-friendly for a wide range of applications.

\item \textbf{Better in Reasoning \& Coding.} The training of \textbf{TeleChat2.5} and \textbf{T1} incorporates reinforcement learning (RL) to explicitly optimize the models' ability to solve mathematical and coding problems, demonstrating substantial improvements compared to their predecessors when tackling complex problems in these domains.

\end{itemize}

In the remainder of this paper, we first present the model architecture in Section \ref{model}. Next, we describe our
pre-training process, including the pre-training recipe, construction of training data, and long context extension techniques in Section \ref{pretraining}. Thereafter, we discuss our post-training methodology, including the composition of training data and specific methods during Supervised Fine-Tuning (Section \ref{sft}), Direction preference optimization (Section \ref{dpo}) and Reinforcement Learning (Section \ref{rl}). We highlight special efforts to improve performance for specific capabilities such as code, math \& reasoning, tool use and precise instruction following in Section \ref{key}. We describe our hardware and infrastructure that powered training and discuss
several optimizations that leads to improvements in training efficiency in Section \ref{engineer}. We then present the detailed evaluation results in Section \ref{result}, covering both the base and chat models.

\section{Model Architecture}
\label{model}
\textbf{TeleChat2}, \textbf{TeleChat2.5}, and \textbf{T1} share a unified model architecture, largely retaining the design of their predecessor, TeleChat. This architecture incorporates a Pre-Norm design with RMSNorm normalization \citep{rmsnorm}, employs SwiGLU \citep{shazeer2020glu} as the activation function for the Feed-Forward Network (FFN), and integrates Rotary Positional Embeddings \citep{su2022roformer}. Detailed network specifications can be
found in Table \ref{architect}. There are several minor adjustments compared to TeleChat, which are detailed below:

\begin{itemize}

\item \textbf{Grouped Query Attention (GQA).} We use Grouped Query Attention with 8 key-value heads instead of the traditional Multi-Head Attention(MHA) for models with 115 billion parameters, achieving both accelerated training and efficient KV cache utilization.

\item \textbf{RoPE base frenquency.} By increasing the RoPE base frequency hyperparameter,  we improve our capacity to handle longer contexts more effectively. See Section \ref{long} for Details.

\end{itemize}

\begin{table}[htbp] \small
\begin{center}
\begin{tabular}{ccccccc}
\toprule
     $Params$ & $n_{layers}$ & $d_{model}$  & $d_{ffn}$ & $n_{heads}$ & $n_{kv\_heads}$ & $n_{vocab}$ \\ 
\midrule
 35B & 64 & 6144 & 20480 & 48 &48 &131072 \\
 115B & 96& 8192 & 40960 & 64& 8& 131072 \\

\bottomrule
\end{tabular}
\caption{Detailed model architecture hyperparamters of TeleChat2, TeleChat2.5 and T1 model family.}\
\label{architect}
\end{center}

\vspace{-2mm}
\end{table}


\section{Pre-training}
\label{pretraining}

\subsection{Overall Pre-training Recipe}
\begin{figure}[htbp] 
\centering 
\includegraphics[width=\textwidth]{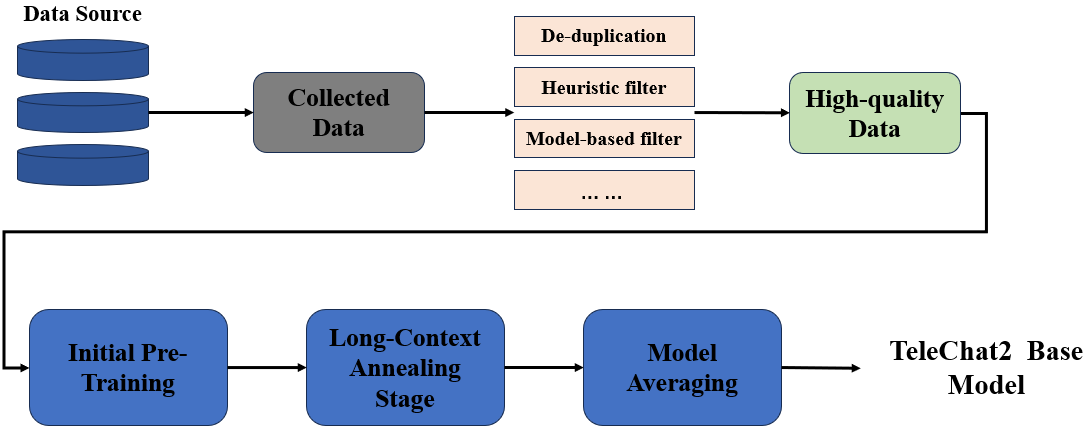} 
\caption{The pre-training framework.} 
\label{pretrain} 
\end{figure}

Our training process for TeleBase2 comprises two main stages. First, during the Initial Pre-training Stage (Section \ref{init}), we curate high-quality training data using filtering and data mixture, resulting in a total of 10 trillion tokens. In this stage, the model acquires foundational language structures and accumulates extensive world knowledge from textual data. Second, during the Long-Context Annealing Stage (Section \ref{long}), we refine the model's capabilities through curated and synthetic datasets, particularly enhancing performance on reasoning and knowledge-based tasks. Simultaneously, we extend the model's context length to 256K tokens. In subsequent sections, we will elaborate on these stages from both data composition and training methodology perspectives. The pre-training framework is illustrated in Figure \ref{pretrain}.

\subsection{Initial Pre-training Stage}
\label{init}

In the initial pre-training phase, our primary goal is to equip the model with broad and comprehensive world knowledge. To achieve this, we train the model on an extensive, high-quality and diverse corpus. The pretraining dataset is meticulously curated and filtered to ensure linguistic richness, topical diversity, and coverage across languages and writing styles. The model is trained on 10 trillion tokens, enabling it to develop a robust understanding of language structures, factual knowledge, and common-sense reasoning, which lays a solid foundation for subsequent finetuning and domain-specific adaptation.

\subsubsection{Data Collection}
Compared to its predecessor, TeleBase2 utilizes a more extensive and higher-quality training dataset during pre-training. We build the pre-training corpus by aggregating diverse, high-quality data from multiple sources to create a robust knowledge foundation. The dataset comprises general-domain and domain-specific content. General-domain sources include web pages, books, encyclopedias, news articles, social media platforms, academic papers, code repositories, and other resources. Domain-specific data is curated from over twenty specialized industries, such as finance, construction, healthcare, and other technical fields.



\subsubsection{Data Cleaning}
Data cleaning is critical to improving model performance by ensuring the quality, consistency, and relevance of training data. We implement a suite of data cleaning and quality assurance techniques, which are outlined below:

\textbf{De-duplication.} We implement a hierarchical de-duplication strategy comprising \emph{URL-level}, \emph{document-level}, and \emph{paragraph-level} de-duplication. Following a similar approach to TeleChat \citep{telechat}, this multi-tiered framework ensures the removal of redundant data while preserving the diversity and quality of the training corpus.

\textbf{Heuristic filtering.} 
We devise several heuristic approaches to enhance the overall quality of the data. Some of the heuristic rules are listed as follows. (1) We exclude texts that are exceedingly brief or lack of substantial informational content. (2) We filter out texts exhibiting an anomalously high or low frequency of punctuation marks. (3) Texts containing an excessive number of dirty words are excluded from the dataset. (4) The code-related data is processed using evaluation criteria specific to the source website. For instance, GitHub project code with a low number of repository stars are excluded from the dataset.

\textbf{Model-based quality filtering.} We integrate large language models (LLMs) into our data filtering pipeline to strengthen quality control. After an initial automated filtering step, we deploy LLMs to conduct in-depth semantic analysis. These models evaluate the text’s relevance, coherence, and fluency while identifying and flagging potentially toxic, biased, or inappropriate content. Additionally, they detect nuanced issues such as logical inconsistencies, off-topic segments, and unnatural language patterns that might not be captured by rule-based systems.

\textbf{Math and Code Data Cleaning.} For mathematical and code-related data, we prioritize correctness and executability during quality assurance. To ensure syntactic correctness, we employ automated scripts and static analysis tools to filter out erroneous data. Code samples are then verified using code execution feedback, while mathematical problems are verified using symbolic computation tools to confirm the accuracy of equations and solutions. We also integrate large language models (LLMs) into our validation workflow. Specifically, LLMs are tasked with reviewing code logic, identifying potential errors, and generating expected outputs for comparison with original data. For mathematical content, models independently solve problems and cross-reference their results with provided solutions. This hybrid approach enables efficient detection of subtle errors that traditional rule-based methods might overlook. Finally, a subset of the data undergoes manual review by human experts to ensure clarity, completeness, and relevance. This includes verifying that code samples are self-contained and well-documented, while mathematical content adheres to standard notation and formatting conventions.

\subsubsection{Determine data composition}
Data composition has a significant impact on model performance. However, for very large models like TeleBase2-115B, it is not feasible to do extensive data composition tuning. To address this problem, we conduct a series of experiments on smaller models (3B and 7B) to evaluate the effect of data mix on model performance. For example, we test varying proportions of Chinese and English corpora in training and observe that the English corpus proportion should not be excessively reduced. This finding can be attributed to two factors: (1) the inherently higher linguistic complexity of Chinese, and (2) the generally lower quality of Chinese corpora compared to English. Based on these insights, we predict the performance of larger models under different data compositions and select the most promising mix for scaling up training.
 

During model training, we adopt a curriculum learning strategy to dynamically adjust the proportions of different data types. In the initial training phases, we emphasize simpler and more general data to help the model establish a strong foundation in language understanding and basic reasoning. As training progressed, we gradually increase the proportion of more complex and specialized data, such as mathematical problems and code-related tasks, allowing the model to incrementally build advanced capabilities. To ensure balanced learning, we conduct comprehensive evaluations every 100 billion tokens using a diverse set of benchmarks or validation set covering all major data types. Based on the evaluation results, we adjust the data sampling ratios in the subsequent training stages, increasing the representation of data types where the model shows relative weakness. This dynamic adjustment process enables the model to maintain steady improvements across all domains, resulting in a more robust and versatile language model.



\subsubsection{Training Details}

We employ the Adam optimizer to pre-train TeleBase2, with the following hyperparameter settings: $\beta_1 = 0.9, \beta_2 = 0.95, \epsilon = 1 \times 10^{-8}$. Cosine learning rate scheduler is used to regulate the learning rate, with the peak learning rate scaled according to the model size. After reaching its maximum value following the warm-up steps, the learning rate gradually decays to 10\% of the peak rate. Weight decay with a factor of $0.01$ is applied to all model parameters except for bias terms. Gradient clipping is enforced with a norm of $1.0$. All learnable parameters are initialized using a normal distribution with a standard deviation of $0.006$. Further hyperparameter configurations are detailed in Table \ref{pretrain_detail_table}. Following the methodology of TeleChat1, we concatenate data from the same source without applying cross-sample attention masking. We set the maximum sequence length to 8K during the first-stage pre-training, and pre-train TeleBase2 on 10T tokens. 

\begin{table}[h!]
\renewcommand{\arraystretch}{1.5}
    \centering
    \begin{tabular}{ccc}
         \hline
         \textbf{HyperParams} &\textbf{TeleBase2-35B} &\textbf{TeleBase2-115B}   \\
         \hline 
         
         \textbf{Peak lr} &$3\times 10^{-4}$ &$2\times 10^{-4}$  \\
         
         \textbf{Batch tokens} &8M &6M  \\

         \textbf{Warm-up fraction} &0.001 &0.001  \\

         \textbf{Rms Norm Epsilon} &$1 \times 10^{-5}$ &$1 \times 10^{-5}$  \\

         \hline
         
    \end{tabular}
    \caption{The hyperparameter details during the pretraining stage of TeleBase2-35B and TeleBase2-115B.}
    \label{pretrain_detail_table}
\end{table}



\subsection{Long-Context Annealing Stage}
\label{long}
To optimize the balance between training efficiency and effectiveness, we integrate long-context extension during the annealing phase. This approach introduces a unified training stage designed to simultaneously improve both general capabilities and long-context understanding in the base model. Specifically, we extend the context window to 256K tokens for TeleBase2-35B and 128K tokens for TeleBase2-115B, while maintaining general capability parity with their 8K-token counterparts. This integration ensures that the model retains strong foundational skills while adapting to extended context requirements.

\subsubsection{data curation}

The training data is categorized into five distinct length intervals: 0–8K, 8K–16K, 16K–32K, 32K–128K, and 128K+. Within each interval, the data is further subdivided by domain (e.g., exams, web pages, code, and other categories) to enable fine-grained analysis. During the annealing phase, the 0–8K interval is combined with other intervals at a 7:3 ratio, prioritizing shorter sequences while gradually introducing longer contexts. Simultaneously, high-quality data from important domains (e.g., exams and code) is upsampled across all length intervals, ensuring robust coverage of critical knowledge sources. This structured approach aligns with principles of data engineering for long-context training \citep{data_engineering_for_long_context}.

\subsubsection{Training Details}
\begin{figure}[htbp] 
\centering 
\includegraphics[width=\textwidth]{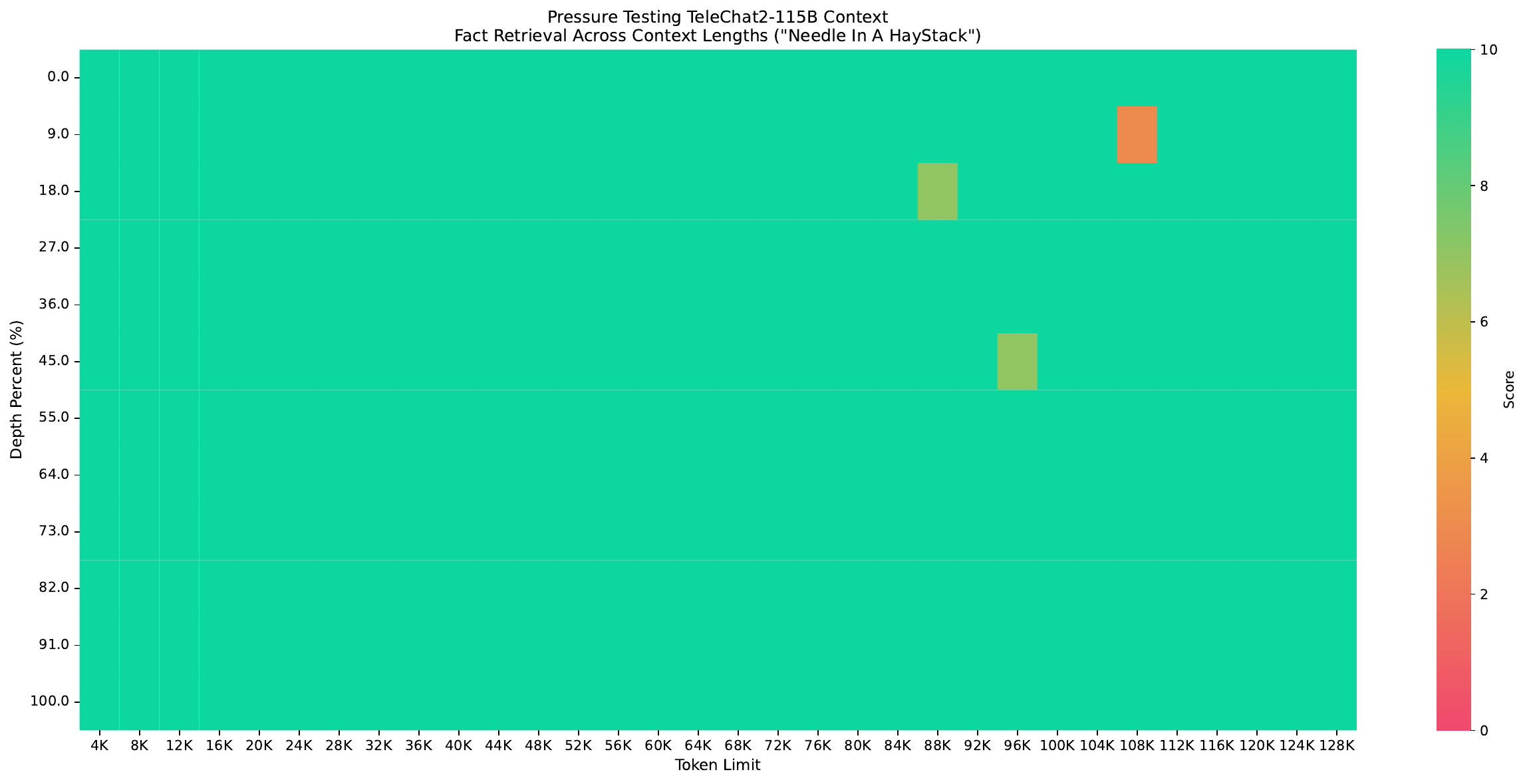} 
\caption{Evaluation results of TeleBase2-115B on the “Needle In A Haystack” test.} 
\label{fig:niah-evaluation} 
\end{figure}

The context length of the pre-trained model is extended sequentially in stages, with the learning rate decreasing successively according to cosine annealing. The initial learning rate for the first annealing stage is equivalent to the learning rate employed during 8K pre-training. Subsequent annealing progresses based on the weights from the 1/3 steps of the preceding training stage, with the learning rate at that time serving as the initial value. As the base in Rotary Position Encoding (RoPE) is a pivotal factor in determining the effective context length of a LLM \citep{liu2024scaling}; \citep{xu2024base}, we set the RoPE's base to $1\times 10^{6}$ for 32K annealing, $8\times 10^6$ for 128K and$4 \times 10^7$ for 256K. Moreover, 50B training tokens are sufficient for each complete annealing phase. After multiple stages of context extension annealing and fine-tuning, the TeleBase2 famliy of models perform well on the “Needle In A Haystack” (NIAH) test over 4K to 128K context lengths. Figure \ref{fig:niah-evaluation} illustrates the evaluation results of TeleBase2-115B.

\subsection{Model Averaging}
To enhance the robustness and generalization of the final model, we apply checkpoint averaging after the training process. Specifically, we compute the element-wise average of parameters of the last five checkpoints. By averaging these checkpoints, we effectively smooth the parameter distribution and improve model stability. 

\subsection{Tokenizer}
\label{tokenizer}
For our tokenizer, we implement BPE with byte-level fallback in SentencePiece, splitting numbers into individual digits as in the approach described by \citep{llama2}. We augment the final vocabulary with special tokens to differentiate dialogue roles and to support tool functionality. To ensure computational efficiency during training and to reserve space for any additional special tokens that might be needed in the future, we configure the model’s vocabulary size to $131072$. We establish a unified vocabulary across all \textbf{TeleChat2}, \textbf{TeleChat2.5} and \textbf{T1} model family, enhancing consistency and reducing potential compatibility issues.

\section{Post-Training}
\begin{figure}[htbp]
    \centering
    \includegraphics[width=1.0\textwidth]{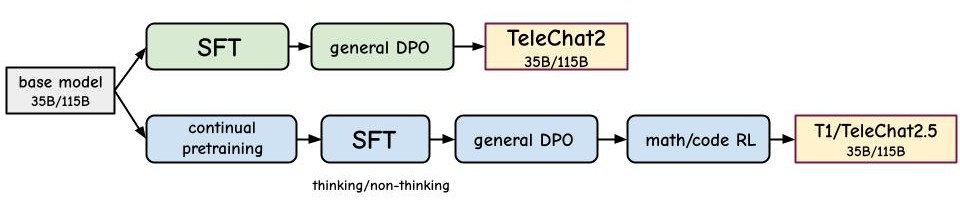} 
    \caption{The development pipelines of TeleChat2, TeleChat2.5 and T1. }
    \label{fig:flow}
\end{figure}

As illustrated in Figure \ref{fig:flow}, the \textbf{TeleChat2} model is trained directly on the base model through a supervised fine-tuning (SFT) stage and a direct preference optimization (DPO) stage to enhance its general capabilities. On the other hand, the \textbf{TeleChat2.5} and \textbf{T1} models first undergo a continual pretraining stage, followed by a three-stage post-training process. This process comprises: (1) an SFT stage with both thinking and non-thinking modes, (2) a DPO stage to improve general capabilities, and (3) a reinforcement learning (RL) stage to strengthen math and coding abilities. This pipeline yields \textbf{T1} (thinking variant) and \textbf{TeleChat2.5} (non-thinking variant).

\subsection{Supervised Fine-Tuning}
\label{sft}
We finetune the pretrained model using high-quality, domain-diverse data including mathematics, coding, reasoning, conversation, model identity, safety, etc. To create high-quality SFT data, we develop a two-stage pipeline comprising (1) diverse query collection and (2) response generation with quality verification.

\subsubsection{Query Collection}

To systematically organize and classify our SFT data, we develop a tagging system that categorizes prompts by domain and discipline, ensuring balanced representation across diverse subject areas. This hierarchical system includes major categories such as mathematics, coding, reasoning, conversational safety, instruction following, tool use, and more. Each category is further subdivided into granular classifications to comprehensively capture the required capabilities.

\textbf{Sourcing from Public Datasets.}  We source queries from a wide array of open-source datasets and employ rigorous data-cleaning processes to eliminate duplicates or highly similar entries. To identify semantic relationships, we map the queries into a high-dimensional embedding space and applied the K-means clustering algorithm to group them effectively.

\textbf{Enhancing Dataset Diversity and Balance.} After cleaning and organizing the data within our tagging system, we identify gaps in certain categories and observe uneven task difficulty distributions. To address these challenges, we utilize self-instruct and instruct-evolution techniques to generate synthetic queries. These methods allow us to construct a query set that not only fully covers knowledge system but also achieves a well-balanced distribution of complexity and diversity. Specifically, we design separate difficulty-scoring prompts for different data categories and leverage LLMs to individually score each data type within every source. For domains such as mathematics and code, we employ a pass rate metric to distinguish the learning difficulty. Regarding certain data types (e.g., creative writing, role-playing, instruction-following, and structured data generation), we observe generally low difficulty levels in open-source datasets. To address this, we manually curate high-quality seed examples and reconstruct datasets through instruction evolution. This approach ensures the data difficulty closely aligns with real-world usage complexity.

\textbf{Query Quality Scoring.}  To guarantee query quality, we implement LLM-based scoring mechanisms. Queries are evaluated against predefined criteria including fluency, standardized formatting, and completeness of contextual information necessary for generating robust responses. Low-scoring queries are either excluded or down-weighted in training to prioritize high-quality data. After thorough review and refinement, we curate a dataset with broad coverage and a well-calibrated difficulty range.

\subsubsection{Response Generation and Quality Control}

We employ both human annotation and synthetic data generation to generate responses.

\textbf{Human annotation collection.} We assemble a team of internal annotators and external contractors to perform manual data annotation. Our annotators possess diverse expertise across a wide range of disciplines. To address queries that challenge current large language models (LLMs), particularly in math and reasoning tasks, we rely on our annotation team to generate high-quality responses. For non-reasoning tasks such as creative writing, role-play, and open-ended QA, we engage human annotators to validate the accuracy of synthetic data.

\textbf{Synthetic data generation.}  For the collected queries, we first generate responses using high-performance models and then select the optimal answer based on task-specific evaluation criteria. Specifically, for tasks with verifiable correctness (e.g., mathematics, code generation, instruction-following, STEM exams), we employ rule-based reward systems to evaluate responses through predefined metrics, and only correct answers are retained. For subjective tasks (e.g., humanities, creative writing, open-ended QA), we utilize LLM-as-judge frameworks, where independent large language models score responses based on fluency, coherence, and relevance. Only the highest-scoring response is retained.

We implement a comprehensive suite of rule-based data verification mechanisms to further ensure data accuracy. The primary rules are listed as follows. (1) During generation, problems including duplicate content, truncated outputs, and illegible characters frequently occur. We strictly filter out such erroneous data. (2) We enforce constraint compliance through rule-based validation scripts, ensuring adherence to format-specific requirements like output length, paragraph count, or structural guidelines imposed by user queries. (3) We implement a content filter using a sensitive keyword database to filter answers potentially containing safety risks. The flagged data is then executed further validation by human annotators for quality assurance.

\subsubsection{Determine Data Mix}

The composition of post-training data critically influences the behavior of language models. To optimize performance, we employ an iterative algorithm that upsamples high-quality data sources while downsampling lower-quality ones in the final data mix. Our analysis reveals a potential negative correlation between model performance and perplexity on the validation set $\mathcal{V}$, which is created by extracting 1\% of the data from our training set $\mathcal{T}$. Specifically, models achieving the better evaluation performance typically exhibit the lower perplexity on validation set. However, when the validation set is partitioned by category, not all subsets reach their minimum perplexity at the same training steps. To address this problem, we designed an algorithm that iteratively adjusts the representation ratio $r_i$ of each category subset $i$ within the overall training data, where $i \in \mathbb{N}^*, 1 \leq i \leq |\mathcal{V}|$. 

During the \textit{t}-th round fine-tuning experiment, we divide the training data into various subsets by their classifications and record the perplexity of each of them at regular training intervals. We set the maximum iteration count to $T$ for termination assurance, $t \in \mathbb{N}, 0 \leq t < T$. Next, we use cubic spline interpolation to fit a curve $p = f_i^{(t)}(s)$, representing the perplexity $p$ of the subset $i$ as a function of the training step $s$ in iteration $t$. Denote the lowest point of this curve as $(s_i^{(t)}, p_i^{(t)})$. Similarly, we compute the weighted average of the perplexities according to the tokens of each subset and fit a curve whose lowest point is denoted as $(\bar{s}^{(t)}, \bar{p}^{(t)})$.

The new proportion can be calculated as follows.

\begin{align}
    r_i^{(t+1)} = r_i^{(t)} \kappa^{\frac{s_i^{t} - \bar{s}^{(t)}}{\mu}},
\end{align}

where $\kappa$ and $\mu$ are hyper-parameters dynamically calibrated based on dataset characteristics, with optimal values of 10 and 15,000 respectively in our experiment.

\begin{align}
    \hat{r}_i^{(t+1)} = \frac{r_i^{(t+1)}}{\sum_{i=1}^{|\mathcal{V}|} r_i^{(t+1)}}.
\end{align}

\begin{figure}[htbp]
    \centering
    \includegraphics[width=0.5\textwidth]{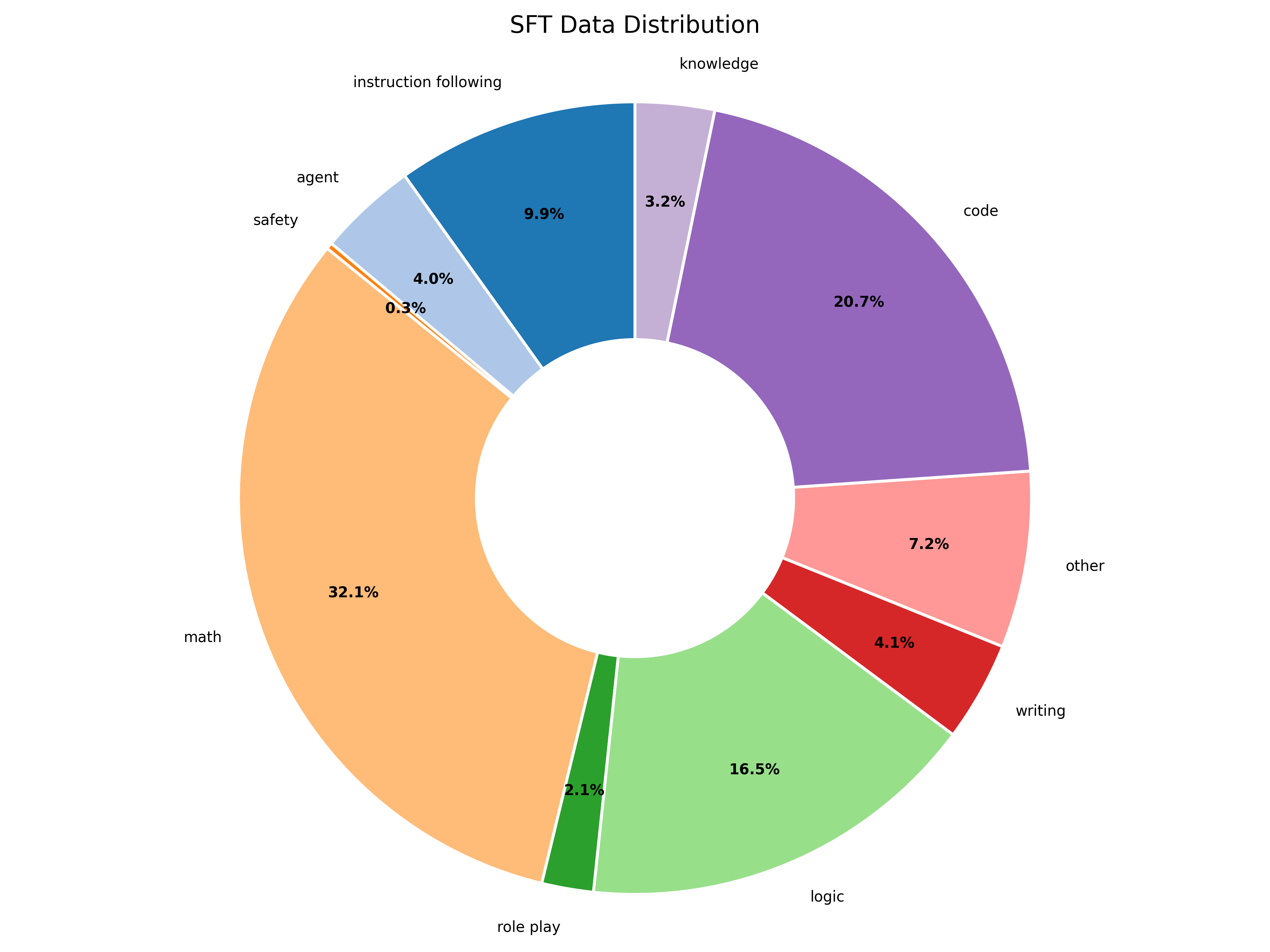} 
    \caption{Data Distribution of Supervised Finetuning data. }
    \label{fig:data_dist}
\end{figure}

After regularization, we apply $\hat{r}_i^{(t+1)}$ as the new proportion in the next round of experiment. The data distribution of supervised finetuning data is demonstrated in Figure \ref{fig:data_dist}.

\subsubsection{Training Details}

We optimize hyperparameters for fine-tuning through grid search, achieving model-specific training configurations. For the 35B variant, the cosine decay learning rate scheduling starts at $3 \times 10^{-5}$ and gradually decays to $1 \times 10^{-5}$ with a batch size of 8; for the 115B variant, the learning rate starts at $1.5 \times 10^{-5}$ and decays to $1.5 \times 10^{-6}$ with a batch size of 16. To enhance training efficiency and reduce sequence padding overhead, we implement a packing strategy that concatenates multiple training samples into a single sequence, while strategically combining single-turn samples into multi-turn dialogues whenever possible, enhancing the model's multi-turn conversational capability. 



\subsection{Direct Preference Optimization}
\label{dpo}

Direct Preference Optimization (DPO, \citep{dpo2023}) is an offline training algorithm designed for learning from preference feedback. It enables direct optimization of the policy using preference data, eliminating the need to construct reward models or sample online from the active policy. The primary objective of DPO is to maximize the margin between the log-likelihood of selected responses and the log-likelihood of rejected responses, while ensuring that the model remains close to the initial policy \citep{ivison2024unpackingdpoppodisentangling}. We apply DPO to align the model with human preferences in general tasks by leveraging pairs of accepted and rejected outputs to discourage undesirable behaviors. In this section, we detail our preference data construction pipeline in section \ref{dpo_data}, and our training details in section \ref{dpo_train}.

\subsubsection{Preference Data Curation}
\label{dpo_data}

The preference data is of utmost importance in improving in the generation quality and performance of large language models. A preference dataset $\mathrm{P}$ typically
comprises prompts, responses, and rankings. Each prompt $x$ is associated with a pair of response $y^+$, $y^-$ (where $y^+$ is the chosen response and $y^-$ is the rejected response), along with a preference ranking between them (indicated as $y^+ \succ y^- | x$). Our process for creating preference data involves four stages: prompt selection, generating responses from a pool of models, annotating preferences using LLM-as-a-judge, and constructing pairs. The specifics of this process are outlined below.

\textbf{Prompt Selection.}
The first step in preparing a dataset for Direct Preference Optimization (DPO) involves the selection of prompts or user instructions for generating responses and gathering preferences. The quality and diversity of the prompt set are essential for ensuring the effectiveness of DPO. Specifically, a high-quality prompt set should fulfill two key criteria: (1) It should demonstrate diversity and cover a wide range of domains, including math \& reasoning, coding, creative writing, and more, to enhance the model’s adaptability and enable it to address a variety of real-world challenges and inquiries. (2) It should encompass a varied mix of easy, moderate, and challenging questions to promote a comprehensive understanding and reduce the risk of overfitting to specific difficulty levels.

We divide the SFT prompts into two parts, allocating 90\% for SFT training and 10\% for DPO. As our SFT prompts offer a well-rounded exposure to diverse domains and different complexity levels, our DPO prompts are able to meet the aforementioned criteria. Additionally, we integrate new instruction-following constraints to enhance the model's ability to adhere to instructions, and also introduce pairs based on the badcases of the previous model to address its weaknesses.

\textbf{Response Generation.}
When given a prompt, we start by sampling from a pool of state-of-the-art open-source and proprietary models, which differ in parameter size and model family. We use greedy sampling and only sample once for each model. Next, we incorporate on-policy data by sampling completions from the latest \textbf{TeleChat2.5} and \textbf{T1} models, utilizing high-temperature sampling to produce multiple responses. To improve the efficiency of rejection sampling, we employ vllm \citep{kwon2023efficientmemorymanagementlarge} to speed up the inference process.

\textbf{Preference Annotation.}
After generating multiple responses for each prompt, it is necessary to assign a reward to each response. (1) For verifiable problems, the reward is determined based on specific criteria or rules. For instance, in coding problems, we evaluate if the solution passes the unit test. In math, reasoning, and standard exam problems, we assess if the generated answer leads to the correct solution. For instruction-following constrained prompts, we verify if the generated answer adheres to the constraints.
(2) For open-ended problems with free-form answers, we use an LLM-as-a-judge \citep{zheng2023judgingllmasajudgemtbenchchatbot} to assess every answer on a scale of 0 to 10 based on four distinct factors: usefulness, adherence to instructions, integrity, and accuracy.

\textbf{Preference Pair Construction.} The construction of preference pairs follows several key principles. 

\begin{itemize}
    \item \textbf{Chosen Responses} are exclusively selected from highest-scoring responses. To maintain response quality standards, we impose a minimum threshold of score $\geq$ 8 for chosen response eligibility. When multiple responses achieve identical maximum scores, priority is given to responses generated by TeleChat series itself rather than off-policy candidates. This design choice mitigates potential distribution shift issues inherent in DPO training, as demonstrated in previous work \citep{dpo2023}.

    \item \textbf{Rejected responses} are strictly sampled from TeleChat series model's own generations. This approach allows the model to self-correct by learning from its own error patterns.

\end{itemize}

A minimum absolute score difference ($\Delta \geq 2$) is enforced between chosen and rejected pairs. This threshold accounts for the documented instability of LLM-as-a-judge scoring, effectively filtering out ambiguous comparisons where minor score variations may not reflect genuine quality differences. For input prompts that generate multiple valid pairs, we randomly sample $K=4$ distinct pairs per input prompt. This results in $98,273$ pairs for DPO training.

\subsubsection{Training Details}
\label{dpo_train}

We train an epoch for DPO, with a learning rate of $5\times 10^{-7}$ and and batch size of $256$. We use a learning rate warm-up and cosine learning rate scheduler. The $\beta$ hyper-parameter is set to be $0.1$. We conduct DPO training on our long-context SFT checkpoints, but only select samples with token length shorter than 8,192. Our observation indicates that utilizing only short-context training data in DPO does not negatively impact long-context performance.

During DPO training, we add an additional negative log-likelihood (NLL) loss term for the pair winners with a scaling coefficient of 0.2, which also proves crucial for performance \citep{pang2024iterativereasoningpreferenceoptimization}. Additionally, we employ a technique of masking out the termination tokens from both selected and rejected responses in the loss function to enhance the stability of DPO training, following a similar approach as described in \citep{grattafiori2024llama3herdmodels}. This is necessary because the existence of shared tokens in both selected and rejected responses creates a conflicting learning objective, requiring the model to simultaneously increase and decrease the likelihood of these tokens.
\subsubsection{Model Merging}
we merge models derived from experiments involving different data versions or hyperparameters during DPO stage. In particular, we merge the multiple models by simply averaging their weights, and observe that this merging process is beneficial for enhancing the model's robustness and overall capabilities.
\subsubsection{Iterative DPO}
The iterative application of the offline preference tuning method procedure has proven beneficial, with the updated model used to construct new preference pairs that are more informative and lead to further improvements. As a result, we apply these methods in three rounds, collecting new preference pairs during each cycle by sampling synthetic data from the latest models.

\subsection{Reinforcement Learning}
\label{rl}
Reinforcement learning (RL) has proven to be effective in enhancing the reasoning capabilities of large language models (LLMs) beyond the Supervised Fine-Tuning (SFT) stage \citep{shao2024deepseekmathpushinglimitsmathematical}. In this work, we focus on optimizing the model's performance in both mathematical reasoning and code generation through reinforcement learning strategies.

\textbf{(1) Mathematical RL.} We curate a dataset from two publicly available sources: OpenR1-Math-220k \footnote{\url{https://huggingface.co/datasets/open-r1/OpenR1-Math-220k}} and Synthetic-1 \footnote{\url{https://huggingface.co/datasets/PrimeIntellect/verifiable-math-problems}}. To ensure data quality, we filter out problems requiring proofs and those with incomplete or inconsistent references. Specifically, we retain only problems that could be automatically verified using the \textit{math\_equal} function\footnote{Available at \url{https://github.com/hkust-nlp/simpleRL-reason/tree/v0}}, which checks numerical or analytical equivalence of answers. For answer extraction, we prompt the model to wrap its final answer in \textit{\\boxed\{\}} and run the verification process to confirm correctness.

\textbf{(2) Coding RL.} We extract a subset of coding problems from the SFT dataset, and only retain samples that are capable of performing code execution feedback. For unit testing, we develop a secure local code sandbox environment supporting diverse testing methods, including standard input-output validation and assertion-based verification.

\textbf{(3) Tool use RL.} We curate the function-call data for reinforcement learning following a two-step strategy:
\textbf{(i) Initial Candidate Set Construction.} We select a batch of function call data originating from the same source as the supervised fine-tuning (SFT) data as candidates. Subsequently, multiple large language models (LLMs) are used to perform multiple inferences on each query. Queries where the outputs are consistent across models, along with their corresponding ground-truth answers, are selected as training inputs.
\textbf{(ii) Difficulty Stratification and Data Curation.} The target model is used to perform multiple inferences on the queries. The model outputs are compared against reference answers to calculate the \texttt{pass@5} rate. Queries are categorized into difficulty levels based on \texttt{pass@5}:
\begin{itemize}
    \item $\texttt{pass@5} = 1$: These queries are too easy for the current model (Easy).
    \item$0< \texttt{pass@5} < 1$: The model has the potential to answer correctly but exhibits unstable performance on these queries (Medium).
    \item $\texttt{pass@5} = 0$: These queries are difficult for the model to answer correctly (Hard).
\end{itemize}
The RL training dataset is composed of medium and hard data in a 2:1 ratio. For the reward function design, we implement category-specific processing based on data type. Specifically, data is divided into tool-requiring and tool-free categories, with the calculation formula as follows:
$$\text{reward} = 
\begin{cases} 
1, & \text{if } I_{\text{tool}}=1 \land M_{\text{format}}=1 \land M_{\text{match}}=1 \\
-1, & \text{if } I_{\text{tool}}=1 \land (M_{\text{format}}=0 \lor M_{\text{match}}=0) \\
2 \times \dfrac{S - S_{\min}}{S_{\max} - S_{\min}} - 1, & \text{if } I_{\text{tool}}=0 
\end{cases}$$

Our reward function design distinguishes based on whether a task requires tool calls. For tasks that do require tool calling, we establish a binary reward: if the model's output format is perfectly correct and the specific content of the tool call exactly matches the reference answer, it receives the full reward ($+1$); if the output format is incorrect or the tool call content deviates from the reference answer, a penalty ($-1$) is given. For pure text tasks that do not require tool calls, we employ a relatively flexible scoring mechanism: First, we use another Large Language Model (LLM) to perform a quality assessment on the model's output, resulting in a raw quality score $S$; Then, we map this raw score onto a unified reward value range of $[-1, 1]$ through a linear transformation formula, in order to enable unified comparison and optimization with the rewards for tool-calling tasks.

We utilize the OpenRLHF \footnote{\url{https://github.com/OpenRLHF/OpenRLHF}} framework for training and employ the \textit{reinforce++} algorithm \citep{hu2025reinforceefficientrlhfalgorithm}. To ensure stable training, we implement dynamic sampling, which continues sampling until the batch is fully filled with examples whose accuracy is neither 0 nor 1, as proposed in \citep{yu2025dapoopensourcellmreinforcement}. For hyperparameters, we use the AdamW optimizer \citep{loshchilov2019decoupledweightdecayregularization} with a constant learning rate of $5\times 10^{-7}$, combined with a linear warm-up over 20 rollout steps. During the rollout phase, the prompt batch size is set to 128, and we generate 16 responses per prompt. For training, the mini-batch size is also configured to 128.

\section{Key Abilities}
\label{key}
We make special efforts to improve performance for specific capabilities including code, reasoning, tool use, long context and precise instruction following.


\subsection{Code}
\textbf{Two-Stage Training Strategy.} We implement a coarse-to-fine two-stage fine-tuning approach. In the first stage, the base model is trained on tens of millions of diverse instruction samples synthesized from large-scale open-source datasets (e.g., CodeAlpaca, CodeSearchNet) and code extracted from GitHub repositories. This foundational phase broadens the model’s capabilities by exposing it to a wide spectrum of tasks. In the subsequent fine-tuning phase, we employ high-quality, meticulously curated instruction datasets. These include multilingual code generation tasks, programming contests (sourced from Codeforces and LeetCode via web crawling), and programming tutorials. For each query, the LLM generates multiple candidate responses. Verifiable problems are evaluated using code execution feedback, while unverifiable problems leverage the LLM itself to rank and select the most suitable example for supervised fine-tuning.

\textbf{Code Execution Feedback.} For problems that support test case verification, we automatically generate 10 test cases using LLMs. These test cases comprehensively cover normal scenarios, boundary conditions, exceptional cases, and complex inputs to rigorously evaluate correctness. The test cases are categorized by programming language (e.g., Python, C, C++, Java, JavaScript) and executed in a secure sandbox environment. Code correctness is validated through runtime execution verification. Samples failing due to errors in code execution (e.g., invalid syntax or assertion error) are filtered out to ensure training data quality.

\textbf{Curriculum Learning.} We implement a model-driven curriculum learning strategy that leverages the model’s own generative capacity to assess prompt difficulty during the second training stage. Specifically, we generate ten responses using a high sampling temperature (e.g., $T=0.6$)  for each prompt. The pass rate (determined by code execution feedback for verifiable tasks) is calculated as a proxy for difficulty, which dynamically construct a training curriculum. Initially, the model focuses on prompts with higher pass rates, ensuring stable learning and foundational skill acquisition. As training progresses, it gradually transits to prompts with lower pass rates, iteratively refining its coding capabilities while systematically expanding its limits.

\subsection{Math and Reasoning}
\textbf{Two-Stage Training Strategy.} For math and reasoning tasks, we adopt a two-stage fine-tuning strategy consistent with code tasks, transitioning from broad capability construction to in-depth precision optimization. In the first stage, the base model is trained on over ten million synthetic samples sourced from extensive open-source datasets (e.g., StackExchange), synthetic K-12 math problems with answers, and synthetic university instructional materials. All data undergoes source quality assessment, deduplication, format cleaning, synthetic data generation, and quality sampling checks. The second stage employs a smaller yet higher-quality curated dataset. Logical reasoning samples are manually collected with ground-truth answers and cover domains such as causal inference, operations research and game theory. Math data includes high-quality open-source datasets (e.g., MATH, GSM8K training sets), licensed K-12 math problems with verified answers, global competition problems (e.g., IMO, AMC), and a small amount of synthetic data to balance distributions. All samples undergo triple verification: problem quality scoring, answer consistency checks, and reasoning process validation. A difficulty-grading mechanism ensures balanced distribution of data across different difficulty levels. 

\textbf{Answer Verification Mechanism.} To validate the accuracy of math answers, we implement a multi-model collaborative verification strategy combined with manual supervision for consensus screening. Specifically, for a target set of math problems, we use multiple large models to independently generate answers. A dedicated answer consistency judgment mechanism analyzes and compares the outputs. Samples with complete agreement across all model outputs proceed to manual sampling quality checks, while inconsistent outputs are re-examined through manual annotation to ensure final answer correctness. 

\subsection{Tool use}
\textbf{Data Curation.} We collect mainstream open-source function call datasets \citep{Zhang2024xLAMAF} \citep{Hermes-Function-Calling-Dataset-V1} \citep{qin2023tool}
\citep{Toshniwal2024OpenMathInstruct1A1}
 \citep{Li2023ModelScopeAgentBY} and perform data cleaning and restructuring. Our validation focuses on two key aspects: 
 \begin{itemize}
     \item Format Validation, where we rigorously check the alignment of the tool calls with the provided function list. This involved verifying: 1) the correct correspondence of tool names, 2) the matching of parameter names, and 3) the compliance of parameter types with requirements. 
     \item Tool Call Result Validation, where we utilize a Large Language Model (LLM) to assess the validity of the tool calls and the accuracy of the tool names and parameter configurations. 
 \end{itemize}
Furthermore, referencing the methodology used in constructing the BFCL benchmark, we categorize the collected function call data to ensure a balanced distribution of function call types within the training dataset.

\textbf{Tool Graph based Data Construction.} After cleaning open-source data, we collect approximately 110K samples. However, during the cleaning process, we identify issues including insufficient Chinese data, limited conversational turns, and low difficulty levels. To address these challenges, we construct a tool-graph structure based on dependency relationships between APIs, leveraging various graph sampling methods to create tasks with balanced difficulty distribution. Furthermore, we utilize the dependency relationships within the tool-graph to facilitate verification of multi-turn tool-calling accuracy, which demonstrates significant optimization effectiveness.

\subsection{Precise instruction following}
To improve the model’s instruction following ability, we develop a systematic pipeline for constructing SFT training datasets. 
In this process, we construct high-quality training data through three key stages: constraint set construction, instruction evolution, and response generation with validation filtering.

\textbf{Constraint Set Construction.}
Following IFEval (\cite{zhou2023instructionfollowingevaluationlargelanguage}), we identify representative application scenarios and construct a constraint set composed entirely of verifiable constraints which can be rigorously validated through automated scripts. For example, these constraints include response length requirements, linguistic norms, formatting guidelines, etc. By leveraging automated validation, this approach eliminates the need for manual intervention.

\textbf{Instruction Evolution.} 
Based on the constraint set, we prompt the LLM to evolve seed instructions into new ones by explicitly incorporating a randomly sampled subset of constraints (typically no more than six). These constraints guide the LLM to generate instructions with clear operational requirements. In addition, the LLM is required to explicitly specify the parameter values corresponding to these constraints (e.g., number of keywords, word limits), which are recorded for subsequent validation.

\textbf{Response Generation with Validation Filtering.} 
Finally, we utilize LLM to generate responses for the newly constructed instructions. Leveraging both the constraint definitions and the parameter values associated with each instruction, we design specialized validation scripts for each type of constraint. These scripts evaluate the model’s outputs based on execution feedback and automatically filter out responses that fail to meet the constraints. This process ensures that the resulting instruction-response pairs consistently adhere to predefined quality standards.

\section{Engineering}
\label{engineer}
We describe the hardware and infrastructure that powered pre-training at scale and describe several optimizations that leads to improvements in training efficiency.

\subsection{Infrastructure}
The previous version of TeleChat\citep{telechat} was trained on a cluster with 640 NVIDIA A100 GPUs. As we scaled up to new series of TeleChat, training was migrated to ctyun’s Shanghai compute center, which provides the computational power essential for training trillion-scale models.

\textbf{Compute.} The new series of TeleChat family were trained on up to Atlas 900 A2 cluster with 8k Ascend NPUs. Each node in the cluster contains 8 HCCS-connected NPUs. Training jobs are scheduled using a MindCluster-based platform.

\textbf{Storage.}  Storage resources comprise Cluster Management (CM) nodes, Metadata Server (MDS) nodes, Object Storage Server (OSS) nodes, and physical storage devices known as OceanDisk.  CM nodes are connected to cloud-based storage systems via dual 25 Gbps links, providing a management interface for distributed storage operations. OceanDisk devices are directly connected to the MDS and OSS nodes using a Fibre Channel (FC) network, ensuring high-speed and low-latency communication for data storage and retrieval. These four types of nodes and devices collectively form the High Performance File System (HPFS) shared storage system, which is optimized for distributed and high-throughput workloads. The HPFS shared storage system is uplinked to the RDMA over Converged Ethernet (RoCE) switches via dual 100GE links, enabling seamless integration with the larger network infrastructure and ensuring high-bandwidth access for compute and storage nodes.

\textbf{Network.} The parameter communication network adopts a two-layer Clos architecture (Charles Clos topology). Each training server connects its 200GE uplink to the RoCE switch, achieving high-speed 200GE RoCE interconnection between processing units. The Spine/Leaf hierarchy is configured with a nonconverged design to ensure maximum bandwidth availability. The parameter communication network incorporates Network-Side Load Balancing (NSLB)  to ensure efficient load balancing at the link layer during large model training. This approach mitigates hash collisions and improves the overall throughput efficiency of the computational cluster.

\subsection{Parallel Computing}
\subsubsection{Parallelism Strategies}
The distributed training of Telechat2 is based on the 4D parallelism strategy provided by MindSpore’s general-purpose large-model parallel framework \citep{mindspore_website}. This framework is designed to support efficient and scalable training of large-scale models by integrating four key parallelism strategies: Data Parallelism (DP; \cite{rajbhandari2020zero}; \cite{zhao2023pytorchfsdpexperiencesscaling}), Tensor Parallelism (TP; \cite{shoeybi2020megatronlmtrainingmultibillionparameter}), Pipeline Parallelism (PP; \cite{huang2019gpipeefficienttraininggiant}; \cite{narayanan2021efficientlargescalelanguagemodel}), and Context Parallelism (CP; \cite{liu2023ringattentionblockwisetransformers}).

\textbf{Data Parallelism (DP):} The input dataset is partitioned along the batch dimension, with different device groups independently processing separate data batches. During backpropagation, gradient synchronization is performed across all devices, ensuring consistent updates to model parameters. This approach is particularly effective for scaling to larger datasets and improving hardware utilization across distributed systems.

\textbf{Tensor Parallelism (TP):} Model weights are partitioned across devices to reduce memory usage and computational overhead. Intermediate results are exchanged and aggregated using collective communication primitives such as All-Gather and ReduceScatter,  which enable efficient distributed computation of tensor operations.

\textbf{Pipeline Parallelism (PP):} The model is divided into layers, or stages, with each stage assigned to a specific group of devices. Forward and backward passes are executed in a pipelined manner to maximize parallelism. To mitigate the inefficiencies caused by pipeline bubbles, strategies such as load balancing and virtual pipeline scheduling are employed.

\textbf{Context Parallelism (CP):} This strategy, unique to MindSpore, implements a 3D sequence parallelism scheme designed to handle long-sequence tasks efficiently. By splitting sequence computations across devices, CP alleviates memory and computation constraints associated with large input sequences.

To determine the optimal parameters for distributed parallelism, we conducted extensive experiments across various configurations. Tensor Parallelism (TP) incurs communication overhead due to operations such as All-Gather and ReduceScatter, while Pipeline Parallelism (PP) is affected by inefficiencies introduced by Bubble and Send/Recv communications. By employing load balancing and other optimization techniques to reduce pipeline bubbles, we found that PP parallelism consistently outperformed TP in terms of efficiency. After carefully tuning the parallelism configuration, hardware resources, and software optimizations, we achieved a Model FLOPs Utilization (MFU; \cite{chowdhery2022palmscalinglanguagemodeling}) of 33.8\%-36.3\% for the configurations
shown in Table \ref{tab:training_config}.

\begin{table}[t]
\caption{Scaling configurations and MFU for TeleChat 2 115B pre-training.}
\label{tab:training_config}
\centering
\small
\begin{tabular}{lrrrrrrrrrr}
\toprule
\textbf{Model Size} & \textbf{Cards} & \textbf{DP} & \textbf{TP} & \textbf{PP} & \textbf{CP} & \textbf{VPP} & \textbf{BS} & \textbf{Seq Len} & \textbf{Tokens/Batch} & \textbf{MFU (\%)} \\
\midrule
115B & 512  & 8 & 8 & 8 & 1 & 1 & 128  & 8k    & 1M  & 36.3 \\
115B & 4096 & 64 & 8 & 8 & 1 & 2 & 512  & 8k    & 4M  & 33.8 \\
115B & 6144 & 96 & 8 & 8 & 1 & 3 & 768  & 8k    & 6M  & 34.1 \\
115B & 4096 & 8 & 8 & 4 & 16 & 2 & 32   & 128k  & 4M  & 34.5 \\
\bottomrule
\end{tabular}
\end{table}
In large-scale distributed training, maintaining precise control over the global batch size is critical for ensuring model convergence and achieving optimal performance. It is well-documented that excessively large batch sizes can adversely affect convergence dynamics and final model quality. For this reason, the global batch size is typically constrained between 4M and 8M tokens during the initial stages of training. When training Telechat-115B on a 4096-NPU cluster, the increased data-parallel (DP) dimension led to a larger tokens per batch. To constrain tokens per batch to 4M, the number of micro-batches in the pipeline was reduced, which increased pipeline bubbles and lowered overall efficiency. To address this, we utilized the Virtual Pipeline Parallelism (VPP) feature to minimize bubbles, resulting in an MFU of 33.8\%. When scaling to a 6144-NPU cluster, we increased the VPP factor to 3, further reducing the pipeline bubble ratio and improving the MFU to 34.1\%. For ultra-long sequence training with a sequence length of 128k, we leveraged Context Parallelism (CP) to alleviate the memory and computational pressure associated with long sequences. This approach enabled training Telechat-115B on a 4096-NPU cluster, achieving an MFU of 34.5\%.

These results demonstrate the effectiveness of carefully balancing parallelism strategies and leveraging advanced features such as VPP and CP to optimize distributed training efficiency, particularly when scaling to large clusters and handling long-sequence datasets.

\subsubsection{Training Optimizations}
In addition to these foundational parallelism strategies, Telechat’s distributed training integrates several advanced optimizations enabled by MindSpore. Selective Re-computation is utilized to reduce memory overhead by recomputing select activations during backpropagation instead of storing them.  Optimizer Parallelism  enhances training efficiency by distributing the computational workload of optimizer operations across devices. Fine-grained multi-replica features allow for overlapping of computation and communication, effectively masking communication latency and improving end-to-end throughput. Furthermore, Pipeline Parallelism Optimizations leverage Virtual Pipeline Parallelism (VPP), employing a 1F1B (one forward, one backward) scheduling strategy combined with pipeline load balancing adjustments to achieve higher utilization of computational resources.

\textbf{Selective Re-computation}. During large-scale model training, activations generated in the forward pass are typically stored for use in the backward pass, resulting in significant memory consumption. This issue is exacerbated in Pipeline Parallelism (PP), where activations from multiple micro-batches must be accumulated, further increasing memory pressure. For models exceeding 70B parameters, a common approach is to omit activation storage and recompute activations during the backward pass, thereby reducing memory usage. However, this method introduces additional computation during backpropagation, potentially lowering computational efficiency.

To address this, the new series of TeleChat training leverages the Selective Re-computation capability provided by MindSpore. This approach selectively applies re-computation to key operators, balancing memory savings with computational overhead. Specifically, we targeted operators within the Feed-Forward Network (FFN), including Silu and Mul, as well as the Cast operator (from fp32 to bf16) in RMSNorm (Root Mean Square Normalization). These operators were chosen for their low computational cost and significant impact on reducing memory allocated for activations. This strategy allowed us to optimize memory usage while maintaining training efficiency.

Additionally, MindSpore supports communication-aware selective re-computation, which, when combined with optimizer parallelism, achieves effects similar to Zero3. MindSpore also enables layer-wise re-computation, selective re-computation, and communication-aware re-computation, further integrated with pipeline parallelism optimizations. These advanced techniques collectively optimize memory allocation and computation, ensuring efficient training of large-scale models.

\textbf{Optimizer Parallelism.} In data-parallel training, parameter updates are redundantly computed across devices, leading to inefficient memory usage and suboptimal performance in large-scale networks. Optimizer Parallelism addresses this issue by distributing optimizer computations across the devices in the data-parallel dimension. Specifically, model parameters and gradients are divided into slices based on device IDs, with each device independently updating its assigned slice. Once updated, the parameters are aggregated across devices using communication operations. This approach offers the benefit of natural load balancing, ensuring that each device has an equal share of parameters and computations. However, it imposes a constraint that parameter shapes must be divisible by the number of devices. The theoretical gains of this method align with parameter sharding, and several optimizations were introduced in TeleChat’s distributed training to enhance its effectiveness.
\begin{itemize}
    \item \textbf{Weight Sharding for Static Memory Reduction:}Model weights are partitioned to further reduce static memory consumption. To preserve the original tensor shapes for forward and backward passes, shared weights are aggregated at the end of each iteration and redistributed before the next iteration’s forward pass.
    \item \textbf{Overlap Communication to Improve Performance:}A primary drawback of optimizer parallelism is the communication overhead associated with sharing weights. By overlapping communication operations with forward computations, we can minimize the perceived communication latency. Specifically, cross-iteration execution of communication allows communication operators to be grouped and fused, enabling efficient interleaving of communication and computation.
\end{itemize}

\textbf{Pipeline Parallelism Optimization.} In pipeline-parallel training scenarios, memory imbalance is a prominent challenge, particularly as the frontend stages often face significant memory pressure. To address this issue, we implemented an optimization strategy that combines adjusting the number of layers assigned to each stage with differentiated recomputation strategies:
\begin{itemize}
    \item \textbf{Memory-Intensive Stages:}For stages experiencing high memory pressure, we reduced the number of layers allocated to these stages and adopted selective recomputation for all layers. This approach maximizes memory savings while balancing computational trade-offs.
    \item \textbf{Memory-Light Stages:}Conversely, stages with less memory pressure were assigned additional layers and employed selective recomputation for only a subset of layers, striking a balance between memory usage and computational efficiency.
\end{itemize}

\begin{figure}[t]
\centering
\includegraphics[width=0.85\textwidth]{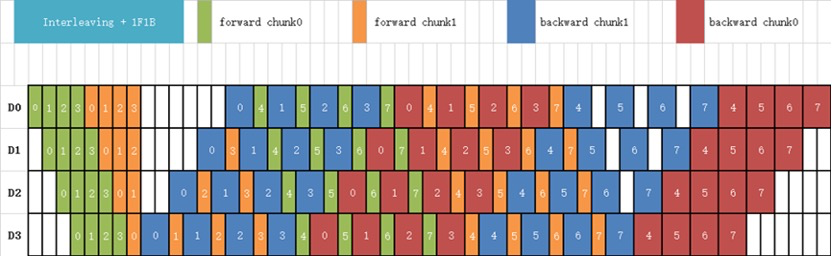}
\caption{Example of Virtual Pipeline Parallelism (VPP) Scheduling: Demonstrates the interleaved computation of non-consecutive layers using the IF1B (One Forward One Backward) strategy. The scheduling mechanism reduces pipeline bubbles by overlapping communication and computation phases, while maintaining load balancing across stages.}
\label{fig:vpp_scheduling}
\end{figure}

To ensure the effectiveness of large-scale model training, batch token sizes are typically constrained (e.g., 8M or 16M). When training with a large cluster, the significant increase in data parallelism (DP) results in a smaller micro-batch size. Under a fixed number of pipeline stages, smaller micro-batches lead to larger pipeline bubbles, negatively impacting training efficiency. To enhance the efficiency of pipeline parallelism and reduce the proportion of bubbles, we adopted Virtual Pipeline Parallelism (VPP) during the training of the TeleChat2 model with 115B parameters. Traditional pipeline parallelism generally assigns consecutive layers (e.g., Transformer layers) to a single stage. In contrast, VPP scheduling employs interleaved computation of non-consecutive layers within each stage(figure \ref{fig:vpp_scheduling}). By increasing communication overhead, this approach significantly reduces the bubble ratio, thereby improving overall training performance.

\textbf{Long-Sequence Optimization}. To support long-sequence training with lengths of 128k–256k tokens, we implemented Sequence Parallelism (also known as Context Parallelism) by splitting the sequence dimension of the query, key, and value (QKV) tensors. This approach effectively reduces memory consumption. During the Attention computation phase, the sequence dimensions of the key and value tensors are reassembled using all-gather communication.

To achieve sequence load balancing, we utilized point-to-point all-gather communication to exchange sequence dimension data of the query and Attention results across devices. This enables swapping computationally intensive sequences from later stages with lighter sequences from earlier stages, ensuring a balanced computation load across devices.

For even longer sequences (e.g., on the order of millions of tokens), we employed the Ring-Attention algorithm provided by MindSpore. This method avoids fully reassembling the sequence dimensions of the key and value tensors during Attention computation. Instead, it performs blockwise computation on local QKV data, ensuring mathematical equivalence while achieving complete load balancing and overlapping computation with communication. This optimization further reduces memory consumption and enhances performance when training on ultra-long sequences.

\subsubsection{Reliability and  Challenges}
During the pretraining phase of TeleChat2, hardware failures were the primary cause of service interruptions, including issues with optical modules, HBM (High Bandwidth Memory), and memory components. In response to these challenges, we implemented the following measures:

\textbf{Recovery Mechanism Optimization}. Optimized failure recovery by improving the mechanisms for storing and loading logs, checkpoints, and data, while upgrading the training framework and scheduling platform. These enhancements significantly reduced the time needed for resuming training after interruptions and preemptively addressed cluster environment issues through version checks.

\textbf{Hardware Reliability Improvements}. Reinforced inspection routines for critical hardware such as HBM, optical modules, and memory. Additionally, stricter standards for hardware replacement were established, and the hardware issue resolution process was streamlined.

\begin{table}[t]
\caption{Hardware Failure Statistics During Pretraining}
\label{tab:failure_stats}
\centering
\small
\begin{tabular}{lrr}
\toprule
\textbf{Failure Category} & \textbf{Count} & \textbf{Proportion (\%)} \\
\midrule
Optical Module Failure & 36 & 19 \\
HBM Failure & 33 & 18 \\
Memory Failure & 14 & 8 \\
DPU Failure & 14 & 8 \\
AI Core Failure & 6 & 3 \\
Optical Cable Failure & 5 & 3 \\
Motherboard Failure & 5 & 3 \\
Hard Drive Failure & 3 & 2 \\
Overheating & 3 & 2 \\
CPU Failure & 2 & 1 \\
NPU Failure & 2 & 1 \\
Power Supply Failure & 2 & 1 \\
RAID Failure & 1 & 1 \\
Controller Failure & 1 & 1 \\
Others & 57 & 31 \\
\bottomrule
\end{tabular}
\end{table}

As a result of these efforts, the weekly failure rate in the mid-to-late stages of pretraining was maintained below 1\%. Training interruptions caused by hardware failures were significantly reduced, with an average Mean Time Between Failures (MTBF) of 4 days and a maximum interval of 21 days for core cluster hardware. The cluster availability metrics were strong, with weekly uptime consistently exceeding 99\% and the longest uninterrupted training session lasting 288 hours.

Despite significant improvements, several critical challenges remain unresolved, which continue to hinder further advancements in training reliability and efficiency. The absence of efficient fault diagnosis tools has resulted in cascading issues, such as difficulty interpreting error codes, challenges in pinpointing the root node of errors, and the inability to monitor shared storage utilization effectively in terms of space and performance. These deficiencies not only prolong downtime during failures but also increase the complexity of troubleshooting and system recovery. Developing robust diagnostic tools and monitoring systems to address these gaps will be essential for minimizing disruptions, optimizing resource utilization, and ensuring seamless scaling in distributed training environments.

\section{Evaluation}
\label{result}

\subsection{Pre-trained Model}

The performance of TeleBase2 is evaluated across a diverse range of benchmarks based on the internal evaluation framework. For base models, the assessment focus on their performance in general knowledge, commonsense, logical reasoning, mathematical problem-solving, and coding capabilities. The benchmarks we evaluated are listed as follows:

\begin{itemize}

\item  \textbf{General knowledge} benchmarks include C-Eval \citep{huang2023ceval} (zero-shot), MMLU \citep{hendrycks2021measuringmassivemultitasklanguage} (5-shot), MMLU-pro \citep{wang2024mmluprorobustchallengingmultitask} (5-shot), CMMLU \citep{li2023cmmlu} (5-shot), GAOKAO \citep{zhang2024evaluatingperformancelargelanguage} (zero-shot), AGIEval \citep{zhong2023agieval} (zero-shot), GPQA \citep{rein2023gpqagraduatelevelgoogleproofqa} (5-shot) and TheoremQA \citep{chen2023theoremqatheoremdrivenquestionanswering} (5-shot). 
\item \textbf{Commonsense} benchmarks include CommonsenseQA \citep{talmor2019commonsenseqaquestionansweringchallenge} (5-shot) and TruthfulQA \citep{lin2022truthfulqameasuringmodelsmimic}  (zero-shot).

\item \textbf{Logical Reasoning} benchmarks include BBH \citep{suzgun2022challengingbigbenchtaskschainofthought} (3-shot) and HellaSwag \citep{zellers2019hellaswagmachinereallyfinish} (zero-shot).

\item \textbf{Mathematical Problem-Solving} benchmarks include GSM8K \citep{hendrycks2021measuringmathematicalproblemsolving} (4-shot), MATH \citep{hendrycksmath2021} (4-shot) and Ape210K \citep{zhao2020ape210klargescaletemplaterichdataset} (1-shot).
\item \textbf{Coding} benchmarks include HumanEval \citep{humanevalchen2021evaluatinglargelanguagemodels} (zero-shot), MBPP \citep{mbppaustin2021programsynthesislargelanguage} (3-shot), Humaneval+ (zero-shot), MBPP+ (3-shot) \citep{liu2023codegeneratedchatgptreally}. 

\end{itemize}

In Table~\ref{tab:result-base}, we compare TeleBase2-35B, trained at context lengths of 8K, 32K, and 256K, with Qwen2.5-32B-base. In Table~\ref{tab:result-base-115}, we compare TeleBase2-115B, trained at context lengths of 8K, 32K, and 128K, with Qwen2.5-72B-base. All models are systematically evaluated using a customized evaluation framework with standardized settings to ensure a fair and rigorous comparison.

\newcolumntype{C}{>{\centering\arraybackslash}X}

\newcolumntype{C}{>{\centering\arraybackslash}X}
\begin{table}[htbp]
\renewcommand{\arraystretch}{1.5}
    \centering
    \caption{Comparison among TeleBase2-35B trained under 8K, 32K and 256K context length and Qwen2.5-32B base model.}
    \label{tab:result-base}
    \begin{tabularx}{\textwidth}{CCCCC} 
        \toprule
        \textbf{Benchmark} & \textbf{TeleBase2-35B-8K} & \textbf{TeleBase2-35B-32K} & \textbf{TeleBase2-35B-256K} & \textbf{Qwen2.5-32B} \\
        \midrule
         \multicolumn{5}{c}{\begin{tabular}{@{}c@{}} \textsl{General Knowledge} \\ \end{tabular}} \\
        \hline
        \textbf{C-Eval} & 87.2 & 87.8 & 86.2 & 86.1 \\
        \textbf{MMLU}  & 72.4 & 74.2 & 71.0  & 75.6\\
        \textbf{MMLU-pro} & 47.0  & 48.4 & 43.0  & 62.1 \\
        \textbf{CMMLU} & 77.2 & 77.9 & 76.7 & 88.3 \\
        \textbf{GAOKAO} & 68.6 & 63.2 & 59.1 & 52.1 \\
        \textbf{AGIEval} & 68.9 & 71.3 & 69.3 & 82.7 \\
        \textbf{GPQA} & 36.5 & 37.8 & 38.0 & 41.5 \\
        \textbf{TheoremQA} & 41.0 & 42.8 & 40.3 & 44.3 \\
        \hline
        \multicolumn{5}{c}{\begin{tabular}{@{}c@{}} \textsl{Commonsense} \\ \end{tabular}} \\
        \hline
        \textbf{CommonsenseQA} &  88.4 & 85.7 & 85.3 & 83.4 \\
        \textbf{TruthfulQA} & 57.2 & 54.0 & 55.0 & 70.0 \\
        \midrule
        \multicolumn{5}{c}{\begin{tabular}{@{}c@{}} \textsl{Logical Reasoning} \\ \end{tabular}} \\
        \hline
         \textbf{BBH} & 81.7 & 82.6 & 82.5 & 70.0 \\
         \textbf{HellaSwag} & 96.2 & 91.6 & 90.2 & 93.0 \\
         \hline
        \multicolumn{5}{c}{\begin{tabular}{@{}c@{}} \textsl{Mathematical Problem-Solving} \\ \end{tabular}} \\
         \hline
          \textbf{GSM8K} & 85.2 & 86.2 & 86.3 & 75.0\\
          \textbf{MATH} & 69.2 & 71.6 & 70.0 & 61.2 \\
          \textbf{Ape210K} & 66.8 & 66.0 & 67.0 & 65.5 \\
        \hline
        \multicolumn{5}{c}{\begin{tabular}{@{}c@{}} \textsl{Coding} \\ \end{tabular}} \\
        \hline
        \textbf{HumanEval} & 73.8 & 70.7 & 73.8 & 78.0\\
        \textbf{MBPP} & 65.2 & 65.2 & 68.9 & 74.0\\
        \textbf{Humaneval+} & 66.0 & 66.0 & 67.4 & 69.5\\
        \textbf{MBPP+} & 70.5 & 71.4 & 70.5 & 70.5\\
        
        \bottomrule
    \end{tabularx}
\end{table}

\newcolumntype{C}{>{\centering\arraybackslash}X}
\begin{table}[htbp]
\renewcommand{\arraystretch}{1.5}
    \centering
    \caption{Comparison among TeleBase2-115B under 8K, 32K and 128K context length and Qwen2.5-72B base model.}
    \label{tab:result-base-115}
    \begin{tabularx}{\textwidth}{CCCCC} 
        \toprule
        \textbf{Benchmark} & \textbf{TeleBase2-115B-8K} & \textbf{TeleBase2-115B-32K} & \textbf{TeleBase2-115B-128K} & \textbf{Qwen2.5-72B} \\
        \midrule
         \multicolumn{5}{c}{\begin{tabular}{@{}c@{}} \textsl{General Knowledge} \\ \end{tabular}} \\
        \hline
        \textbf{C-Eval} & 94.0 & 92.3 & 91.0 & 89.5\\
        \textbf{MMLU}  & 81.0 & 79.9 & 78.9  & 77.2 \\
        \textbf{MMLU-pro} & 53.2  & 53.0 & 52.5  & 63.8 \\
        \textbf{CMMLU} & 82.0 & 81.3 & 80.0 & 90.3 \\
        \textbf{GAOKAO} & 73.6 & 72.3 & 73.7 & 68.9 \\
        \textbf{AGIEval} & 69.7 & 70.0 & 71.8 & 84.7 \\
        \textbf{GPQA} & 41.3 & 41.3 & 38.3 & 40.3  \\
        \textbf{TheoremQA} & 44.8 & 45.8 & 45.3 & 46.5 \\
        \hline
        \multicolumn{5}{c}{\begin{tabular}{@{}c@{}} \textsl{Commonsense} \\ \end{tabular}} \\
        \hline
        \textbf{CommonsenseQA} &  86.7 & 85.3 & 85.7 & 87.1 \\
        \textbf{TruthfulQA} & 62.6 & 61.6 & 61.0 & 71.0 \\
        \midrule
        \multicolumn{5}{c}{\begin{tabular}{@{}c@{}} \textsl{Logical Reasoning} \\ \end{tabular}} \\
        \hline
         \textbf{BBH} & 81.5 & 82.7 & 82.8 & 85.1 \\
         \textbf{HellaSwag} & 97.4 & 92 & 92.6 & 96.8 \\
         \hline
        \multicolumn{5}{c}{\begin{tabular}{@{}c@{}} \textsl{Mathematical Problem-Solving} \\ \end{tabular}} \\
         \hline
          \textbf{GSM8K} & 90.3 & 84.5 & 86.0 & 76.5 \\
          \textbf{MATH} & 72.0 & 74.0 & 72.4 & 62.0 \\
          \textbf{Ape210K} & 68.8 & 72.0 & 67.7 & 66.5 \\
        \hline
        \multicolumn{5}{c}{\begin{tabular}{@{}c@{}} \textsl{Coding} \\ \end{tabular}} \\
        \hline
        \textbf{HumanEval} & 72.6 & 69.5 & 67.7 & 78.7 \\
        \textbf{MBPP} & 70.0 & 69.4 & 68.0 & 75.2 \\
        \textbf{Humaneval+} &  65.9 & 63.4 & 61.6 & 71.3 \\
        \textbf{MBPP+} & 71.0 & 71.9 & 68.8 & 71.4 \\
        
        \bottomrule
    \end{tabularx}
\end{table}

\subsection{Post-trained Model}
To comprehensively evaluate the quality of instruction-tuned models, we utilized automated benchmarking frameworks to assess performance of the thinking model (\textbf{T1}) and the non-thinking model (\textbf{TeleChat2} and \textbf{TeleChat2.5}). The instruct models are evaluated under the following benchmarks to compare their capabilities.

\begin{itemize}
\item \textbf{AlignBench} \citep{liu2024alignbenchbenchmarkingchinesealignment} is a comprehensive, multi-dimensional benchmark for evaluating Chinese large language models (LLMs) in alignment with human values and real-world requirements. It comprises 8 core categories, 683 real-scenario queries, and human-verified references.

\item \textbf{IFEval} \citep{zhou2023instructionfollowingevaluationlargelanguage} is a benchmark that evaluates large language models’ ability to follow verifiable instructions. It provides 25 instruction types and around 500 prompts, each with quantifiable criteria. 

\item \textbf{BFCL} (Berkeley Function-Calling Leaderboard) \citep{patil2025bfcl} is a benchmark designed to evaluate large language models' (LLMs) function calling and tool use capabilities. The benchmark employs multi-dimensional evaluation methodologies, including single-turn function calling, multi-turn function calling, and hallucination detection. The BFCL benchmark results presented in this paper specifically reflect the single-turn performance on python-ast track, reporting averages for both non-live and live subtasks.

\item \textbf{MATH500} is derived from the original MATH dataset \citep{hendrycksmath2021}, which comprises 5K mathematical problems.

\end{itemize}

For \textbf{T1} models, we employ a sampling temperature of 0.6, top-p of 0.95, top-k of 50, and repetition penalty of 1.05. For \textbf{TeleChat2} and \textbf{TeleChat2.5}, models use greedy search with a repetition penalty of 1.01. For both modes, we set the max output length to 32,768 tokens. The evaluation results for TeleChat model series alongside comparisons with other models with comparable parameter sizes under similar settings, are presented in Tables \ref{tab:result-sft-35} and \ref{tab:result-sft-115}.

The evaluation results demonstrate TeleChat series models' robust capabilities across both thinking and non-thinking modes. \textbf{T1-115B} achieves exceptional performance in thinking mode, surpassing OpenAI o1-mini by +4.0 points on MATH500 (94.0 vs. 90.0) and +0.31 on Alignbench (8.22 vs. 7.91). In non-thinking mode, \textbf{TeleChat2.5-115B} outperforms GPT-4o-1120 by +12.0 points on MATH500 (87.0 vs. 75.0) and demonstrates a +4.74 advantage in BFCL (83.39 vs. 78.65). The TeleChat2.5-35B variant also remains competitive against similarly sized alternatives. Compared to Deepseek-R1-Qwen32B-distill, \textbf{TeleChat2.5-35B} achieves +5.67 points on IFEval (78.26 vs. 73.33) and +3.97 points on BFCL (80.11 vs. 76.14), demonstrating stronger performance in thinking mode. 

\newcolumntype{C}{>{\centering\arraybackslash}X}
\begin{table}[htbp]
\renewcommand{\arraystretch}{1.5}
    \centering
    \caption{Comparison among \textbf{T1-35B}, \textbf{TeleChat2-35B},\textbf{TeleChat2.5-35B} and other models under thinking/non-thinking mode with comparable parameter sizes.}
    \label{tab:result-sft-35}
    \begin{tabular}{ccccc} 
        \toprule
        \textbf{Benchmark} & \textbf{MATH500} & \textbf{Alignbench} & \textbf{IFEval} & \textbf{BFCL} \\
        \hline
        \multicolumn{5}{c}{\begin{tabular}{@{}c@{}} \textsl{Thinking} \\ \end{tabular}} \\
        \hline
        \textbf{T1-35B} & 90.0 & 7.93 &78.26 &80.11 \\
        \textbf{Deepseek-R1-Qwen32B-distill}  & 94.3 & 7.42 & 73.33  & 76.14 \\
        \textbf{QWQ-32B} & 96.0  & 7.97 & 80.09  & 83.10\\
        \textbf{Qwen3-32B} &  93.0 & 8.27 & 85.92 & 86.82 \\
        \hline
        \multicolumn{5}{c}{\begin{tabular}{@{}c@{}} \textsl{Non-Thinking} \\ \end{tabular}} \\
        \hline
        \textbf{TeleChat2-35B} & 61.0 & 6.97 & 77.74 & 75.32 \\
        \textbf{TeleChat2.5-35B} & 77.0 & 7.74 &78.52 &78.28 \\
        \textbf{Qwen2.5-32B} & 82.0 & 7.39&79.44 &82.11 \\
        \textbf{Qwen3-32B(non-thinking)} & 83.0 & 8.23 & 84.07 & 81.84 \\
        
        \bottomrule
    \end{tabular}
\end{table}

\newcolumntype{C}{>{\centering\arraybackslash}X}
\begin{table}[htbp]
\renewcommand{\arraystretch}{1.5}
    \centering
    \caption{Comparison among \textbf{T1-115B}, \textbf{TeleChat2.5-115B}, \textbf{TeleChat2-115B} and other models under thinking/non-thinking mode. }
    \label{tab:result-sft-115}
    \begin{tabularx}{\textwidth}{ccCXXX} 
        \toprule
        \textbf{Benchmark} & \textbf{\#Params}&\textbf{MATH500} & \textbf{Alignbench} & \textbf{IFEval} & \textbf{BFCL} \\
        \hline
        \multicolumn{6}{c}{\begin{tabular}{@{}c@{}} \textsl{Thinking} \\ \end{tabular}} \\
        \hline
        \textbf{T1-115B} & 115B & 94.0 & 8.22 &80.15 &83.39 \\
        \textbf{OpenAI o1-mini} & Unknown & 90.0 & 7.91 & 79.07  & - \\
        \textbf{Deepseek-R1}& 671B(A37B) & 97.2  & 8.43 & 83.70  & 88.68 \\
        \hline
        \multicolumn{6}{c}{\begin{tabular}{@{}c@{}} \textsl{Non-Thinking} \\ \end{tabular}} \\
        \hline
        \textbf{TeleChat2-115B} & 115B & 72.0 & 7.76 & 79.25 & 77.47 \\
        \textbf{TeleChat2.5-115B} & 115B & 87.0 & 7.94 & 80.93 &83.39 \\
        \textbf{Qwen2.5-72B}& 72B &82.0 & 7.62 & 83.70 & 79.15 \\
        \textbf{GPT-4o-1120} & Unknown & 75.0 & 7.49 & 80.18 & 78.65 \\
        \textbf{Deepseek-V3} & 671B(A37B) & 90.2 & 8.06 & 86.10 & 77.66 \\
        
        \bottomrule
    \end{tabularx}
\end{table}

\section{Conclusion}

The introduction of \textbf{TeleChat2}, \textbf{TeleChat2.5}, and \textbf{T1} series represents a significant advancement in large language model (LLM) development. Despite minimal architectural changes, these models achieve substantial performance improvements through systematic upgrades in both pre-training and post-training stages. By publicly releasing these models with scalable parameter configurations (35B and 115B), we empower researchers and developers to leverage cutting-edge LLMs for diverse applications, fostering innovation in natural language processing, code generation and reasoning. This work not only addresses critical gaps in prior research but also provides a robust foundation for future studies on large-scale model optimization and task-specific adaptation.

\section*{Acknowledgements}

We are grateful to the MindSpore Team and SAC (Small \& Agile Commando) Team at Huawei for their engineering support in developing and optimizing the model on the Ascend cluster infrastructure.

\bibliography{iclr2024_conference}
\bibliographystyle{iclr2024_conference}

\clearpage

\appendix

\end{document}